\documentclass[letterpaper, 10 pt, conference]{ieeeconf}  

\IEEEoverridecommandlockouts                              

\newlength{\wdth}

\makeatletter
\newcommand*{\rom}[1]{\expandafter\@slowromancap\romannumeral #1@}
\makeatother

\usepackage{graphics} 
\usepackage{graphicx}
\usepackage{epsfig} 
\usepackage{epstopdf} 
\usepackage{amsmath} 
\usepackage{amssymb}  

\usepackage{microtype}
\usepackage[usenames,dvipsnames,table]{xcolor}
\usepackage{commath}
\usepackage{todonotes}
\usepackage{bm}
\usepackage{mathtools}

\usepackage[utf8]{inputenc}
\usepackage[english]{babel}
\PassOptionsToPackage{hyphens}{url}\usepackage[colorlinks, citecolor={blue}, linkcolor={blue}]{hyperref}
\usepackage{import}
\usepackage{paralist}
\usepackage{algpseudocode}
\usepackage{algorithm}
\usepackage{gensymb}
\usepackage{dsfont}
\usepackage{siunitx}
\usepackage[bottom]{footmisc}
\usepackage{framed}
\usepackage{balance} 
\usepackage{caption}
\usepackage[skins]{tcolorbox}  
\usepackage{subcaption}

\usepackage{todonotes}

\DeclareCaptionFont{mysize}{\fontsize{8}{10}\selectfont}
\usepackage{pifont}

\usepackage{colortbl}	
\usepackage{xspace}

\usepackage{mathrsfs}
\usepackage{siunitx}

\newcommand\I[1]{\mathbf{I}_{#1}}
\newcommand\Zero[1]{\mathbf{0}_{#1}}
\newcommand\Zeros[2]{\mathbf{0}_{#1 \times #2}}

\newcommand\X{\mathbf{X}}
\newcommand\Y{\mathbf{Y}}
\newcommand\Z{\mathbf{Z}}

\newcommand\A{\mathbf{A}}
\newcommand\B{\mathbf{B}}

\newcommand\uVec{\mathbf{u}}
\newcommand\vVec{\mathbf{v}}

\newcommand\J{\mathbf{J}}

\newcommand\R{\mathbb{R}}

\newcommand\ghat[2]{\hm{\left[} #2 \hm{\right]}^\wedge_{#1}}

\newcommand\gexp[1]{\exp_{#1}}

\newcommand\gexphat[1]{\exp_{#1}^\wedge}

\newcommand\glogvee[1]{\log_{#1}^\vee}
\newcommand\gadj[1]{\textup{Ad}_{#1}}

\newcommand\gljac[1]{\mathbf{J}^l_{#1}}

\newcommand\SO[1]{\textup{SO}(#1)}
\newcommand\so[1]{\mathfrak{so}(#1)}
\newcommand\SE[1]{\textup{SE}(#1)}

\newcommand\SEk[2]{\textup{SE}_{#1}(#2)}

\newcommand\T[1]{\textup{T}(#1)}

\newcommand\Exp{\textup{Exp}}

\newcommand\yRot[2]{{\prescript{{#1}}{}{\mathbf{\tilde{R}}}_{#2}}}

\newcommand\yGyro[2]{{\prescript{{#2}}{}{\tilde{\omega}}_{#1, #2}}}

\newcommand\noiseGyro[1]{{\mathbf{w}_{g}^{#1}}}

\newcommand\noiseLinVel[1]{{\mathbf{w}_{v}^{#1}}}
\newcommand\noiseAngVel[1]{{\mathbf{w}_{\omega}^{#1}}}

\newcommand\fkNoiseLinVel[1]{{\mathbf{n}_{v}^{#1}}}
\newcommand\fkNoiseAngVel[1]{{\mathbf{n}_{\omega}^{#1}}}

\newcommand\encoders{\mathbf{\tilde{s}}}
\newcommand\encoderSpeeds{\mathbf{\dot{\tilde{s}}}}
\newcommand\encoderNoise{\mathbf{w}_\mathbf{s}}

\newcommand{\cov}{\mathbf{P}}
\newcommand{\Q}{\mathbf{Q}}

\newcommand{\Xhat}{\mathbf{\hat{X}}}
\newcommand{\err}{\epsilon}
\newcommand{\errG}{\eta}

\newcommand\kpred{{k+1 \mid k}}

\newcommand\kcurr{{k}}
\newcommand\knext{{k+1}}

\newcommand{\expectation}[1]{\mathbb{E}\hm{[} {#1} \hm{]}}

\newcommand\Transform[2]{{\prescript{{#1}}{}{\mathbf{H}}_{#2}}}

\newcommand\Rot[2]{{\prescript{{#1}}{}{\mathbf{R}}_{#2}}}
\newcommand\Pos[2]{{\prescript{{#1}}{}{\mathbf{o}}_{#2}}}
\newcommand\PosBar[2]{{\prescript{{#1}}{}{\mathbf{\bar{o}}}_{#2}}}
\newcommand\oDot[2]{{\prescript{{#1}}{}{\mathbf{\dot{o}}}_{#2}}}

\newcommand\relativeJacobianLeftTrivLinIn[2]{{\prescript{{#2}}{}{\mathbf{S}}^{\text{lin}}_{#2, #1}}}
\newcommand\relativeJacobianLeftTrivAngIn[2]{{\prescript{{#2}}{}{\mathbf{S}}^{\text{ang}}_{#2, #1}}}
\newcommand\relativeJacobianLeftTrivIn[2]{{\prescript{{#2}}{}{\mathbf{S}}_{#2, #1}}}

\newcommand\relativeJacobianLeftTrivAng[2]{{\prescript{{#2}}{}{\mathbf{S}}^{\text{ang}}_{#1, #2}}}

\newcommand\twistMixedTriv[2]{{\prescript{{#2[#1]}}{}{\textbf{v}}_{#1, #2}}}

\newcommand\omegaLeftTriv[2]{{\prescript{{#2}}{}{\omega}_{#1, #2}}}
\newcommand\omegaRightTriv[2]{{\prescript{{#1}}{}{\omega}_{#1, #2}}}

\newcommand\jointPos{\mathbf{{s}}}
\newcommand\jointVel{\mathbf{\dot{s}}}

\newcommand\force[1]{{}_{{#1}} \bm{f}}
\newcommand\wrenchExt[1]{{}_{{#1}} \mathbf{f}^{\mathbf{x}}}
\newcommand\forceExt[1]{{}_{{#1}} \bm{f}^{\mathbf{x}}}
\newcommand\torqueExt[1]{{}_{{#1}} {\tau}^{\mathbf{x}}}

\title{\LARGE \bf
Whole-Body Human Kinematics Estimation using Dynamical Inverse Kinematics and Contact-Aided Lie Group Kalman Filter
}

\author{Prashanth Ramadoss$^{1, 2}$, Lorenzo Rapetti$^{1, 3}$, Yeshasvi Tirupachuri$^{1}$, Riccardo Grieco$^{1}$,  \\
Gianluca Milani$^{1}$, Enrico Valli$^{1}$, Stefano Dafarra$^{1}$,  Silvio Traversaro$^{1}$, and Daniele Pucci$^{1}$ 
\thanks{$^{1}$ Artificial and Mechanical Intelligence, Italian Institute of Technology,
Genoa, Italy, {\tt\small (e-mail: name.surname@iit.it)}}
\thanks{$^{2}$ DIBRIS, University of Genoa, Genoa, Italy}
\thanks{$^{3}$ University of Manchester, Manchester, United Kingdom}
}

\begin{document}

\maketitle
\thispagestyle{empty}
\pagestyle{empty}


\begin{abstract}
Full body motion estimation of a human through wearable sensing technologies is challenging in the absence of position sensors.
This paper contributes to the development of a model-based whole-body kinematics estimation algorithm using wearable distributed inertial and force-torque sensing.
This is done by extending the existing dynamical optimization-based Inverse Kinematics (IK) approach for joint state estimation, in cascade,  to include a center of pressure based contact detector and a contact-aided Kalman filter on Lie groups for floating base pose estimation.
The proposed method is tested in an experimental scenario where a human equipped with a sensorized suit and shoes performs walking motions.
The proposed method is demonstrated to obtain a reliable reconstruction of the whole-body human motion. \looseness=-1
\end{abstract}

\setlength{\belowdisplayskip}{3pt} \setlength{\belowdisplayshortskip}{3pt}
\setlength{\abovedisplayskip}{3pt} \setlength{\abovedisplayshortskip}{3pt}
\allowdisplaybreaks

\section{Introduction}
\par
Several Human-Robot Collaboration (HRC) scenarios require the realization of reliable motion tracking algorithms for the human.
A common approach relies on the use of wearable sensing technologies to achieve the kinematic reconstruction.
A reliable reconstruction becomes challenging in the absence of position sensors when using only proprioceptive measurements.
This paper contributes towards the development of whole body joint state estimation and floating base estimation using only distributed inertial and force-torque sensing by cascading a Inverse Kinematics (IK) approach with a contact-aided filtering on matrix Lie groups.\looseness=-1
\par
In the literature, sensor fusion of measurements from distributed Inertial Measurement Units (IMUs) has been used to estimate human motion assuming bio-mechanical models for the human body and coarsely known location of these sensors in the highly articulated kinematic chain of the model. 
\cite{miezal2016inertial} compares Extended Kalman Filter (EKF)-based and optimization-based sensor fusion methods for inertial body tracking accounting for robustness to model calibration errors and absence of magnetometer measurements for yaw corrections. The authors further extend their work to ground contact estimation and improve lower-body kinematics estimation in \cite{miezal2017real}. 
A human moving in space is usually modeled as a floating base system which implies the necessity to consider the evolution of the configuration space of internal joint angles and base pose over a differentiable manifold.
\cite{joukov2017human} and \cite{sy2019estimating} both exploit the theory of Kalman filtering over matrix Lie groups for motion estimation by explicitly considering the non-Euclidean geometry of the state space, demonstrating significant improvements over other state representations while maintaining the required computational efficiency for real-time applications. \looseness=-1
\cite{von2017sparse} and \cite{sy2019estimating} both propose human motion tracking with reduced IMU counts while exploiting the complex geometry of the underlying configuration space of the human model and integrating additional physical constraints. 
The former aims for a full body motion reconstruction while the latter only focuses on the lower body kinematics. 
Recently, \cite{rapetti2020model} proposed a model-based approach referred to as Dynamical Optimization-based IK, combining dynamical systems’ theory with the IK
optimization, for human motion estimation and
demonstrated full-motion reconstruction on a human model with 48-DoFs and a 66-DoFs in real-time.
However, most of the methods described above either rely on position sensors for full-body estimation or only perform partial kinematics estimation, such as lower body kinematics. \looseness=-1

\begin{figure}[!t]
\centering
\includegraphics[scale=0.15, width =0.32\textwidth]{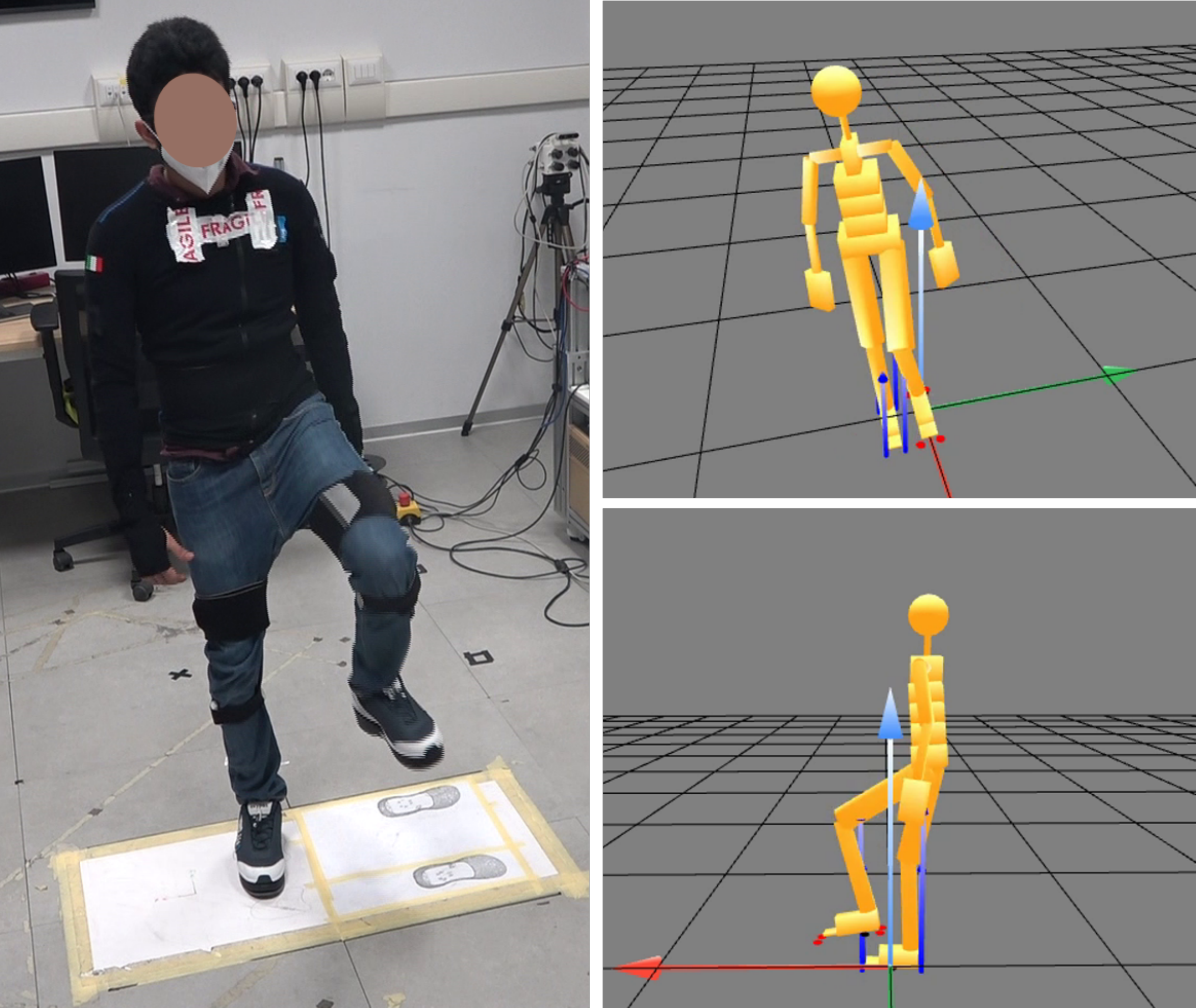}
\caption{Reconstruction of walking-in-place motion using the proposed method.}
\vspace{-2em}
\label{fig:walk-in-place}
\end{figure}

\par
In this paper, we present an approach for whole-body kinematics estimation in the absence of position sensors by extending the joint state estimation from the Dynamical IK \cite{rapetti2020model} with a floating base estimation using a  Center of Pressure (CoP) based contact detection and an EKF on Lie groups through a cascading architecture.
This decomposition is done to handle the floating base estimation within a prediction-correction framework and be able to incorporate contact-aided kinematic measurements without the need for imposing  additional constraints in the IK, while remaining computationally light-weight when applied to systems with high degrees of freedom, such as an articulated human model. 
The validation of the proposed method is demonstrated by reconstructing walking-in-place (Figure \ref{fig:walk-in-place}) and free-walking motions performed by a human subject wearing a sensorized suit and shoes.\looseness=-1

This paper is organized as follows. Section \ref{sec:BACKGROUND} introduces the mathematical notation. 
Section \ref{sec:ESTIMATION} describes the problem of human motion estimation in the absence of position sensors and describes the overall architecture of the proposed approach. 
Sections \ref{sec:DYN-IK}, \ref{sec:CONTACT_DETECTION}, and \ref{sec:EKF} each describe the main blocks of the proposed architecture: Dynamical IK, contact detection and the EKF on Lie groups respectively.
This is followed by experimental evaluation of the proposed method in Section \ref{sec:RESULTS} and concluding remarks in Section \ref{sec:CONCLUSION}. \looseness=-1

\section{Mathematical Background}
\label{sec:BACKGROUND}

\par

\subsection{Notations and definitions}
\subsubsection{Coordinate Systems and Kinematics}
\begin{itemize}
	
	\item ${C[D]}$ denotes a frame with origin $\Pos{}{C}$ and orientation of coordinate frame ${D}$;
	\item $\Pos{A}{B}\in \R^3$ and $\Rot{A}{B}\in \SO{3}$ are the position and orientation of a frame $B$ with respect to the frame $A$;
	
	\item given $\Pos{A}{C}$ and $\Pos{B}{C}$,  $\Pos{A}{C} = \Rot{A}{B} \Pos{B}{C} + \Pos{A}{B}= \Transform{A}{B} \PosBar{B}{C}$, where $\Transform{A}{B} \in \SE{3}$ is the homogeneous transformation and $\PosBar{B}{C} = \begin{bmatrix}\Pos{B}{C}^T & \ 1\end{bmatrix}^T \in \R^4$ is the homogeneous representation of $\Pos{B}{C}$; 
	
	\item $\twistMixedTriv{A}{B} = \oDot{A}{B} = \frac{d}{dt}(\Pos{A}{B}) \in \R^3$ denotes the linear part of a mixed-trivialized velocity \cite[Section 5.2]{traversaro2019multibody} between the frame $B$ and the frame $A$ expressed in the frame ${B[A]}$. $\omegaRightTriv{A}{B} \in \R^3$ denotes the angular velocity between the frame ${B}$ and the frame ${A}$ expressed in ${A}$; \looseness=-1
	\item ${A}$ denotes an absolute or an inertial frame; ${B}$, ${LF}$, and ${RF}$ indicate the frames attached to base link, left foot and right foot respectively; \looseness=-1
\end{itemize}

\subsubsection{Lie Groups}
We refer the readers to \cite[Section II]{ramadoss2021diligentkio} for a detailed description on the notation used for Lie groups. \looseness=-1

\subsubsection{Miscellaneous}
\begin{itemize}
\item $\I{n}$ and $\Zero{n}$ denote the $n \times n$ identity and zero matrices;

\item Given $\uVec, \vVec \in \R^3$ the \emph{hat operator} for $\SO{3}$ is $S(.): \R^3 \to \so{3}$, where $\so{3}$ is the set of skew-symmetric matrices and $S(\uVec) \; \vVec = \uVec \times \vVec$;  $\times$ is the cross product operator in $\R^3$. \looseness=-1

\end{itemize}

\section{Problem Statement}
\label{sec:ESTIMATION}
This section describes the problem of motion estimation for a human equipped with a sensorized suit of distributed IMUs and shoes with embedded force-torque sensors. The human is modeled as an articulated multi-rigid body system. \looseness=-1

The internal configuration of an articulated mechanical system is usually estimated through inverse kinematics with the help of task space or target measurements.
Consider a set of $n_p$ frames $P =\{P_1, P_2, \dots, P_{N_p}\}$ associated with target position measurements $\Pos{A}{{P_i}}(t) \in \R^3$ and target linear velocity measurements $\oDot{A}{P_i}(t) \in \R^3$.
Similarly, consider a set of $n_o$ frames $O = \{O_1, O_2, \dots, O_{N_o}\}$ associated with target orientations $\Rot{A}{{O_j}}(t) \in \SO{3}$ and target angular velocity measurements $\omegaRightTriv{A}{O_j}(t) \in \R^3$.
Given the kinematic description of the model, IK is used to find the state configuration $(\mathbf{q}(t), \nu(t))$, \looseness=-1
\begin{gather}
\scalebox{0.91}{
\setcounter{MaxMatrixCols}{20}
$\begin{aligned}
\label{chap:human-motion:ik-problem}
    \begin{cases}
        \Pos{A}{{P_i}}(t) = h^p_{P_i}(\mathbf{q}(t)), & \forall\ i = 1, \dots, n_p \\
        \Rot{A}{{O_j}}(t) = h^o_{O_j}(\mathbf{q}(t)), & \forall\ j = 1, \dots, n_o \\
        \oDot{A}{{P_i}}(t) = \mathbf{J}^\text{lin}_{P_i}(\mathbf{q}(t)) \nu(t), & \forall\ i = 1, \dots, n_p \\
        \omegaRightTriv{A}{{O_j}}(t) = \mathbf{J}^\text{ang}_{O_j}(\mathbf{q}(t)) \nu(t), & \forall\ j = 1, \dots, n_o \\
        \A^\mathbf{q}\ \jointPos(t) \leq \mathbf{b}^\mathbf{q}, \\
        \A^\nu\ \jointVel(t) \leq \mathbf{b}^\nu,
    \end{cases}
\end{aligned}$
}
\end{gather}
where, $\mathbf{q}(t) \triangleq \mathbf{q}^B(t)$ and $\nu(t) \triangleq \nu^{B}(t)$ is the position and velocity of the mechanical system, $h^p_{P_i}(\mathbf{q}(t))$ and $h^o_{O_j}(\mathbf{q}(t))$ are the position and orientation selection functions from the forward kinematics of frames $P_i$ and $O_j$ respectively,  while $\mathbf{J}^\text{lin}_{P_i}(\mathbf{q}(t))$ and $\mathbf{J}^\text{ang}_{O_j}(\mathbf{q}(t))$ are the linear and angular part of the Jacobian matrix mapping system velocity to target velocities.
$\A^\mathbf{q}$ and $\mathbf{b}^\mathbf{q}$ are constant parameters representing the limits for joint configuration $\jointPos(t)$, and $\A^\nu$ and $\mathbf{b}^\nu$ represent the limits for joints velocity $\jointVel(t)$.

We account for target measurements provided by sparsely distributed IMUs across the body.
Consider IMUs $S_i$ attached to links $L_i$ of the body.
The target orientations $\Rot{A}{{L_i}}$ can be expressed using the absolute orientation measurements $\yRot{{A_{S_i}}}{{S_i}}$ from the IMU as $\Rot{A}{{L_i}} = \Rot{A}{{A_{S_i}}} \ \yRot{{A_{S_i}}}{{S_i}} \ \Rot{{S_i}}{{L_i}}$ and the target angular velocities can be expressed using the gyroscope measurements as $\omegaRightTriv{A}{{L_i}} = \Rot{A}{{L_i}} \yGyro{A}{{L_i}}$, where $\Rot{{S_i}}{{L_i}}$ defines the rotation of the sensor link's frame with respect to sensor frame and $\Rot{A}{{A_{S_i}}}$ is a calibration matrix used to express the measurements of sensor $S_i$ in a common reference frame $A$.  
The matrices $\Rot{{S_i}}{{L_i}}$ and $\Rot{A}{{A_{S_i}}}$ are assumed to be known from a prior calibration procedure. \looseness=-1

The set of target measurements can be collected in a pose target tuple $\mathbf{x}(t)$ and a velocity target vector $\mathbf{v}(t)$,
\begin{gather}
\scalebox{0.91}{
\setcounter{MaxMatrixCols}{20}
$\begin{aligned}
\label{eq:tuple-collection}
\mathbf{x}(t) \triangleq 
\begin{pmatrix} 
\Pos{A}{{B}}(t) \\ \Rot{A}{{B}}(t) \\ \Rot{A}{{L_1}}(t) \\ \vdots \\ \Rot{A}{{L_{n_L}}}(t)
\end{pmatrix}, \quad
\mathbf{v}(t) \triangleq 
\begin{bmatrix} 
\oDot{A}{{B}}(t) \\ \omegaRightTriv{A}{{B}}(t) \\  \omegaRightTriv{A}{{L_1}}(t) \\ \vdots \\ \omegaRightTriv{A}{{L_{n_L}}}(t)
\end{bmatrix}.
\end{aligned}$
}
\end{gather}
It must be noted that the base position $\Pos{A}{{B}}(t)$ and linear velocity $\oDot{A}{{B}}(t)$ are not directly measured but are passed as quantities estimated from the previous time-instant.
Similar to Eq. \eqref{eq:tuple-collection}, the forward kinematics map and the Jacobians necessary for the differential kinematics can be stacked as, \looseness=-1
\begin{gather}
\scalebox{0.91}{
\setcounter{MaxMatrixCols}{20}
$\begin{aligned}
h(\mathbf{q}(t)) \triangleq 
\begin{pmatrix} 
h^p_{P_1}(\mathbf{q}(t)) \\ \vdots \\ h^p_{P_{n_p}}(\mathbf{q}(t)) \\ h^o_{O_1}(\mathbf{q}(t)) \\ \vdots \\ h^o_{O_{n_o}}(\mathbf{q}(t))
\end{pmatrix}, \quad
\mathbf{J}(\mathbf{q}(t)) \triangleq 
\begin{bmatrix} 
\mathbf{J}^\text{lin}_{P_1}(\mathbf{q}(t)) \\ \vdots \\ \mathbf{J}^\text{lin}_{P_{n_p}}(\mathbf{q}(t)) \\ \mathbf{J}^\text{ang}_{O_1}(\mathbf{q}(t)) \\ \vdots \\ \mathbf{J}^\text{ang}_{O_{n_o}}(\mathbf{q}(t))
\end{bmatrix},
\end{aligned}$
}
\end{gather}
leading to the set of equations,
\begin{equation}
\label{chap:human-motion:state-configuration}
\begin{split}
    &\mathbf{x}(t) = h(\mathbf{q}(t)) \\
    &\mathbf{v}(t) = \mathbf{J}(\mathbf{q}(t)) \nu(t).
\end{split}
\end{equation}

The problem of motion estimation then requires to solve for the floating base pose $\Transform{A}{B}$ and velocity $\twistMixedTriv{A}{B}$ of the human along with the internal joints configuration $\jointPos$, and velocity $\jointVel$ using a set of absolute target orientations $\Rot{A}{{L_i}}$ and target angular velocities $\omegaRightTriv{A}{{L_i}}$ obtained from each of the sparsely distributed IMUs along with contact-aided kinematic corrections using contact wrench measurements obtained from the sensorized shoes and a known human model. \looseness=-1

\subsection{Proposed Architecture}
\begin{figure*}[t!]
    \centering
    \includegraphics[width=0.8\textwidth, height=75mm]{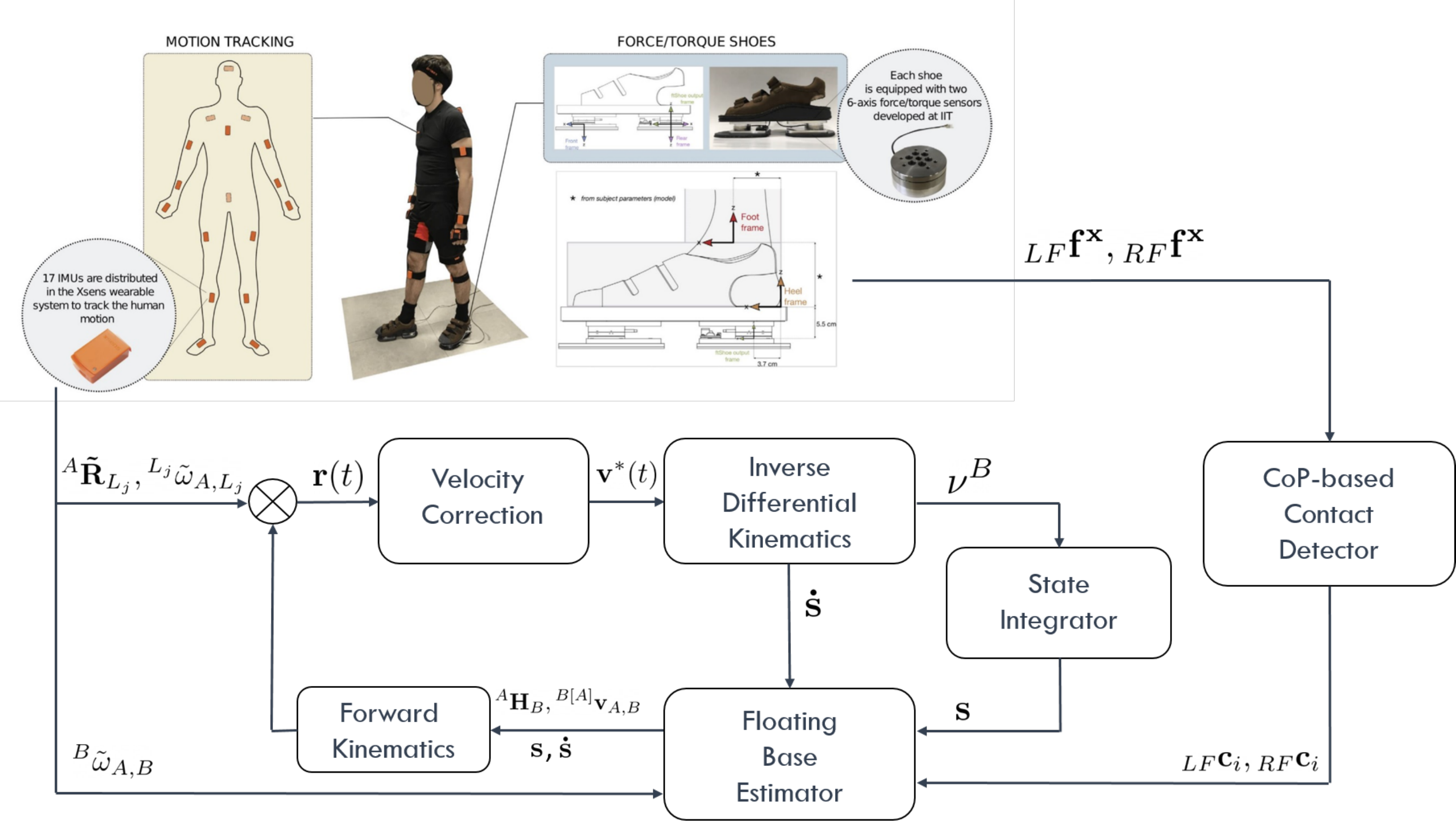}
    \caption{Overall system architecture. Distributed IMU measurements are passed through the sub-blocks of dynamical IK to obtain joint states, while contact wrench measurements are used by the contact detector to obtain contact states. The outputs of these blocks are then passed to the base estimator along with the base gyroscope measurement for obtaining the base pose estimates.}
    \vspace{-1.5em}
    \label{fig:sys_architecture}
\end{figure*}
We propose a system architecture composed of three macro-blocks to tackle the human motion estimation problem, as shown in Figure \ref{fig:sys_architecture}.
Firstly, a \emph{dynamical optimization based IK} block is used to compute the joint configuration from the target measurements of the distributed IMUs with the knowledge of the human model.
The dynamical optimization approach aims to drive the state configuration $(\mathbf{q}(t), \nu(t))$ such that the forward and differential kinematics of the system converge to the reference target measurements in multiple time steps. \looseness=-1

Meanwhile, a \emph{contact detection} block infers ground contact using wrench measurements from the sensorized shoes.
A rectangular approximation is used for the foot geometry to infer the contact state of candidate contact points on the foot based on the local center of pressure (CoP). \looseness=-1

Finally, the outputs of these two blocks are passed to an EKF block. 
The EKF is formulated as a filter over matrix Lie groups \cite{bourmaud2013discrete, barrau2016invariant} employing right-invariant, left-invariant, and non-invariant observations in order to estimate the floating base pose and velocity along with the position of the candidate contact points with respect to an inertial frame.
The EKF block is activated after the dynamical IK block has converged below a certain error threshold for a reliable state initialization, which is crucial for filter convergence.
The estimated base state may or may not be subsequently fed back to the IK block. \looseness=-1

\section{Dynamical Inverse Kinematics}
\label{sec:DYN-IK}
The overall structure of Dynamical IK consists of three main steps, where at first the measured velocity $\mathbf{v}(t)$ is corrected using a forward kinematics error $\mathbf{r}(t)$ to produce an updated velocity vector $\mathbf{v}^{*}(t)$ which is then passed through the inverse differential kinematics module to compute the system velocity $\nu(t)$. Finally, the velocity $\nu(t)$ is integrated to obtain the system configuration $\mathbf{q}(t)$.
These three steps establish a closed-loop regulator on the model-based forward kinematics, thereby, driving the estimated system configuration to the true configuration reflected by the target measurements.
Due to space constraints, we kindly refer the reader to the original paper \cite{rapetti2020model} for more details about Dynamical IK.
The base state integrated from the dynamical IK block is discarded in favor of the EKF block for base state estimation, while the joint state estimates from the dynamical IK are used as inputs to the EKF. \looseness=-1

\looseness=-1

\section{Center of Pressure based Contact Detection}
\label{sec:CONTACT_DETECTION}
\begin{figure}[!t]
\centering
         \centering
\includegraphics[scale=0.2]{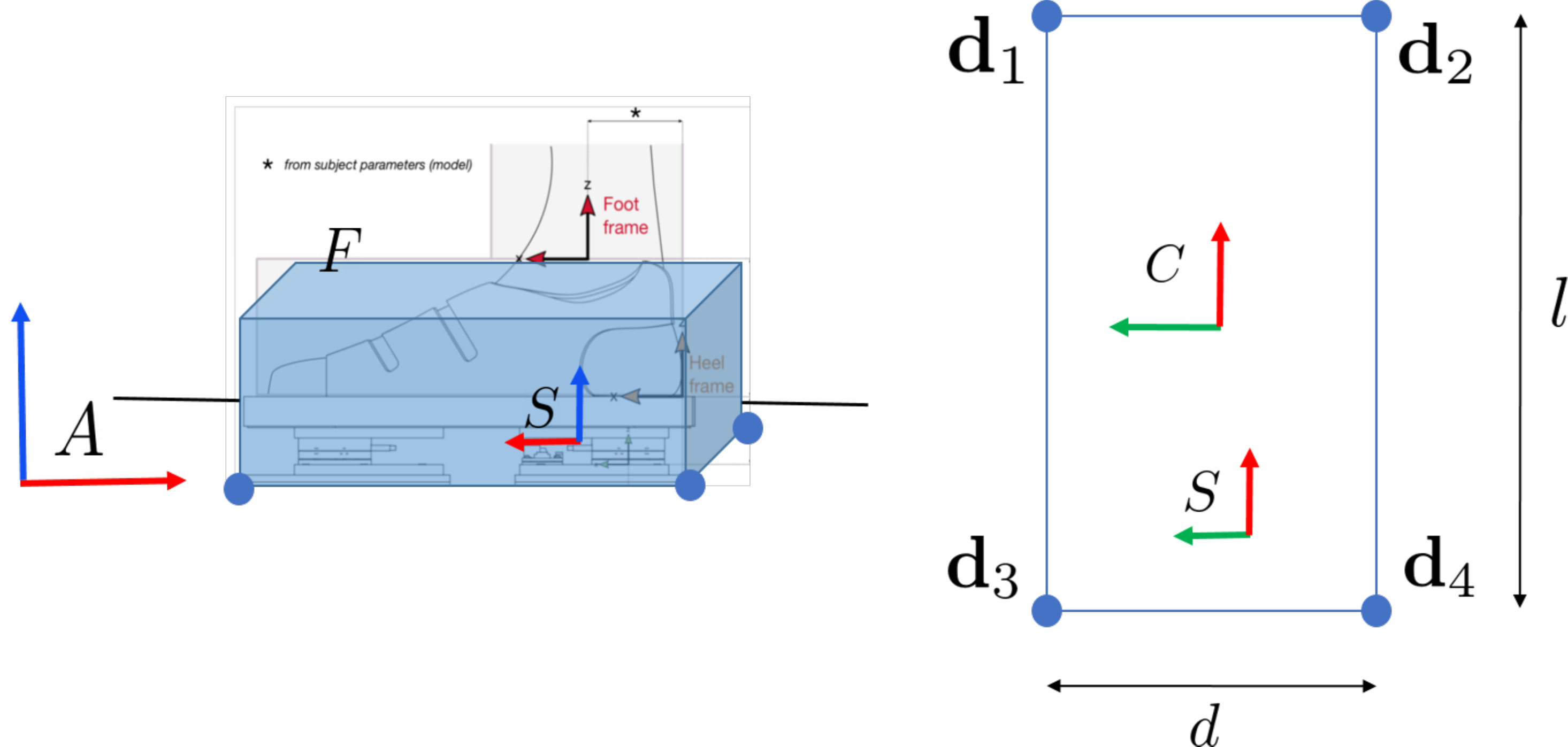}
\caption{Rectangular approximation of the foot geometry with length $l$ and width $d$. $F$, $S$ and $C$ denote the foot, sole and sole-center frames.}
\vspace{-2em}
\label{fig:foot-geometery}
\end{figure}

The contact-aided floating base estimation relies on the contact states of candidate points chosen from the vertices of a rectangular approximation of the foot sole, as shown in the Figure \ref{fig:foot-geometery}.
The contact wrench acting on the foot is decomposed into contact normal forces acting at these vertices. A Schmitt Trigger-based thresholding of these normal forces is then used to infer if a vertex is in contact  with the environment. \looseness=-1

The contact normal force decomposition follows the approach presented in \cite[Appendix A]{dafarra2020predictive} based on the local center of pressure of the foot. 
A coordinate frame $C$ is placed at the center of the sole surface, with the $x$-axis pointing forward and parallel to the side with length $l$, while $z$-axis points upwards and perpendicular to the surface of the foot-sole. \looseness=-1

Given length $l$ and width $d$ of the rectangular sole, the coordinates of the vertices in the frame $C$ are $\mathbf{d}_1 = (\frac{l}{2}, \frac{d}{2}, 0), \mathbf{d}_2 = (\frac{l}{2}, -\frac{d}{2}, 0), \mathbf{d}_3 = (-\frac{l}{2}, \frac{d}{2}, 0),$ and $\mathbf{d}_4 = (-\frac{l}{2}, -\frac{d}{2}, 0)$. \looseness=-1
A contact wrench $\wrenchExt{} = \begin{bmatrix}{\forceExt{}}^T & {\torqueExt{}}^T \end{bmatrix}^T \in \R^6$ acting on frame $C$ is related to the 3D forces $\force{}_i \in \R^3,\ i = \{1,2, 3, 4\}$ at the vertices through the following conditions,
\begin{gather}
\begin{aligned}
   & \force{}_1 + \force{}_2 + \force{}_3 + \force{}_4 = \forceExt{}, \\
   & \sum \mathbf{d}_i \times \force{}_i = \torqueExt{}.
\end{aligned}
\end{gather}
From the equations above, the contact normal forces can be extracted to correspond to the total contact normal force and the contact tangential torques applied along $x$- and $y$-axes.
\begin{align}
\label{eq:chap:human-motion:cop-normal-force}
   & {f_1}_z + {f_2}_z + {f_3}_z + {f_4}_z = f^\mathbf{x}_z, \\
\label{eq:chap:human-motion:cop-tau-x}
   & \frac{d}{2}({f_1}_z + {f_3}_z) - \frac{d}{2}({f_2}_z + {f_4}_z) = \tau^\mathbf{x}_x, \\
\label{eq:chap:human-motion:cop-tau-y}   
   & \frac{l}{2}({f_3}_z + {f_4}_z) - \frac{l}{2}({f_1}_z + {f_2}_z) = \tau^\mathbf{x}_y.
\end{align}

The Center of Pressure (CoP) in the frame $C$ is given by,
\begin{align}
    & \text{CoP}_x = - \frac{\tau^\mathbf{x}_y}{f^\mathbf{x}_z},\quad \text{CoP}_y = \frac{\tau^\mathbf{x}_x}{f^\mathbf{x}_z}.
\end{align}
On expressing the contact normal forces as a function of ${f_4}_z$ and imposing a positivity constraint on the contact normal forces, ${f_i}_z \geq 0$, we can define $\alpha_i = {{f_i}_z}/{f^\mathbf{x}_z}$ as the ratio of the normal force at the vertex to the total contact normal force, leading to the inequalities,
\begin{gather}
\scalebox{0.9}{
\setcounter{MaxMatrixCols}{20}
$\begin{aligned}
\label{eq:chap:human-motion:alpha4-ineq}
\begin{matrix}
    \alpha_4 \geq 0, 
    & \alpha_4 \geq - \frac{\text{CoP}_y}{d} - \frac{\text{CoP}_x}{l}, \\[1em]
    \alpha_4 \leq \frac{1}{2} - \frac{\text{CoP}_x}{l}, 
    & \alpha_4 \leq \frac{1}{2} - \frac{\text{CoP}_y}{d}.
\end{matrix}
\end{aligned}$
}
\end{gather}
With the choice of $\alpha_4$ to equal the average of the CoP bounds, we have a suitable solution for the ratios of contact normal force decomposition,
\begin{gather}
\scalebox{0.9}{
\setcounter{MaxMatrixCols}{20}
$\begin{aligned}
\label{eq:chap:human-motion:alpha4-ineq}
\begin{split}
    & \alpha_4 = \frac{\left(  \max\left(0, - \frac{\text{CoP}_y}{d} - \frac{\text{CoP}_x}{l} \right) + \min\left(\frac{1}{2} - \frac{\text{CoP}_y}{d}, \frac{1}{2} - \frac{\text{CoP}_x}{l} \right) \right)}{2}, \\
    & \alpha_1 = \alpha_4 + \frac{\text{CoP}_x}{l} + \frac{\text{CoP}_y}{d}, \\
    & \alpha_2 = -\alpha_4 + \frac{1}{2} - \frac{\text{CoP}_y}{d}, \\
    & \alpha_3 = -\alpha_4 + \frac{1}{2} - \frac{\text{CoP}_x}{l}.
\end{split}
\end{aligned}$
}
\end{gather}
With this choice of ratios, $\alpha_i$ all equal to $1/4$ when the CoP lies in the center of the foot and the ratios are adjusted appropriately as the CoP moves around the rectangular surface of the approximated foot. 
Such a decomposition and a chosen threshold value for vertex contact detection allow the inference of the contact status of candidate points.

\section{Lie Group Extended Kalman Filter}
\label{sec:EKF}
We employ an EKF over matrix Lie groups with right-invariant, left-invariant, and non-invariant observations using contact information and joint configuration.
This block acts replaces the sub-module of the state integrator block that integrates the base link velocity to obtain the base pose. \looseness=-1

\subsection{State Representation}
The EKF block maintains its own internal state and the results of the Dynamical IK block are considered as measurements passed to the filter.
We wish to estimate the position $\Pos{A}{B}$, orientation $\Rot{A}{B}$, linear velocity  $\oDot{A}{B}$, and angular velocity $\omegaLeftTriv{A}{B}$ of the base link $B$ in the inertial frame $A$. 
Additionally we also consider the position of candidate contact points on the feet $\Pos{A}{{F_{1,\dots,n_f}}}$ and orientation of the feet $\Rot{A}{F}$, where $F = \{{LF}, {RF} \}$ in the state. 
The information about each foot incorporated into the internal state of the EKF consists of four vertex positions of the foot ($n_f = 4$) along with a rotation of the foot.

For the sake of readability, we introduce some shorthand notation for this section.
The tuple of state variables $(\Pos{A}{B},\Rot{A}{B}, \oDot{A}{B}, \Pos{A}{{F_{1,\dots,n_f}}})$ or $(\mathbf{p}, \mathbf{R}, \mathbf{v}, \mathbf{d}_{F_{1 \dots n_f}})$ are encapsulated within $\SEk{N+2}{3}$ matrix Lie group \cite{hartley2020contact}. 
The left-trivialized angular velocity of the base link $\omegaLeftTriv{A}{B}$ denoted as $\omega$ forms a translation group $\T{3}$, while feet rotations $\Rot{A}{F}$ or $\mathbf{Z}_F$ evolve over the group of rotations $\SO{3}$. 
Together, a state space element $\X \in \mathcal{M}$ is represented by a composite matrix Lie group, $\SEk{N+2}{3} \times \T{3} \times \SO{3}^2$.
We simplify the derivation in the remainder of this section by considering only one foot with only one contact point per foot and dropping the suffix $F_{1, \dots, {n_f}}$ and $F$ for the feet positions $\mathbf{d}_{F_{1 \dots n_f}}$ and orientation $\mathbf{Z}_F$ respectively. \\
A state space element $\X \in \mathcal{M}$ keeping only one foot contact point and one foot orientation is then given by, \looseness=-1
\begin{gather}
\scalebox{0.82}{
\setcounter{MaxMatrixCols}{20}
$\begin{aligned}[b]
\label{eq:chap:human-motion:ekf-state-repr}
\X_1 =
\begin{bmatrix}
\mathbf{R} & \mathbf{p} & \mathbf{v} & \mathbf{d}   \\
\Zeros{1}{3} & 1 & 0 & 0  \\
\Zeros{1}{3} & 0 & 1 & 0   \\
\Zeros{1}{3} & 0 & 0 & 1   
\end{bmatrix},  \X_2 =
\begin{bmatrix}
\I{3} & \omega    \\
\Zeros{1}{3} & 1 
\end{bmatrix}, \X = \text{blkdiag}(\X_1, \X_2, \Z).
\end{aligned}$
}
\end{gather}
In tuple representation, we can denote the state space element as, $\X \triangleq (\mathbf{p}, \mathbf{R}, \mathbf{v}, \mathbf{d},\omega, \mathbf{Z})_\mathcal{M}$.
The hat operator $\ghat{\mathcal{M}}{\err}:  \R^{9+3{n_f}+3+3} \to \mathfrak{m}$ with $n_f = 1$ and $N = 1$ that transports a vector $\err$ to the Lie algebra $\mathfrak{m}$ is defined as as,
\begin{gather*}
\scalebox{0.88}{
\setcounter{MaxMatrixCols}{20}
$\begin{aligned}[b]
\label{eq:chap:human-motion:ekf-state-hat}
& \ghat{\SEk{N+2}{3}}{\err_1} =
\begin{bmatrix}
S(\err_\mathbf{R}) & \err_\mathbf{p} & \err_\mathbf{v} & \err_\mathbf{d}   \\
\Zeros{1}{3} & 0 & 0 & 0  \\
\Zeros{1}{3} & 0 & 0 & 0   \\
\Zeros{1}{3} & 0 & 0 & 0   
\end{bmatrix}, \quad \ghat{\T{3}}{\err_2} =
\begin{bmatrix}
\Zero{3} & \err_\omega    \\
\Zeros{1}{3} & 0 
\end{bmatrix}, 
\end{aligned}$
}
\end{gather*}
\begin{gather}
\scalebox{0.9}{
\setcounter{MaxMatrixCols}{20}
$\begin{aligned}[b]
 \ghat{\mathcal{M}}{\err} = \text{blkdiag}\left(\ghat{\SEk{N+2}{3}}{\err_1},\  \ghat{\T{3}}{\err_2},\  S(\err_\mathbf{Z})\right),
\end{aligned}$
}
\end{gather}
with, $\err_1 = \text{vec}(\err_\mathbf{p}, \err_\mathbf{R}, \err_\mathbf{v}, \err_\mathbf{d})$ and the vector $\err \triangleq \text{vec}(\err_1, \err_\omega, \err_\mathbf{Z})$. \looseness=-1
The exponential mapping of the state space expressed in a tuple representation is given by, \looseness=-1
\begin{equation}
\label{eq:chap:human-motion:ekf-state-exp-map}
\begin{split}
\gexphat{\mathcal{M}}\left(\err\right) &= \big(\J\left(\err_\mathbf{R}\right)\ \err_\mathbf{p},\ \Exp\left(\err_\mathbf{R}\right),\ \J\left(\err_\mathbf{R}\right)\ \err_\mathbf{v}, \\
& \quad \ \ \ \J\left(\err_{\mathbf{Z}}\right)\ \err_{\mathbf{d}},\ \err_\omega,\ 
\Exp\left(\err_{\mathbf{Z}}\right)\big)_{\mathcal{M}}.
\end{split}
\end{equation}
where, $\Exp(.) \triangleq\ \gexphat{\SO{3}}(.)$ is the exponential map of $\SO{3}$ and $\J(.) \triangleq \gljac{\SO{3}}(.)$ is the left Jacobian of $\SO{3}$ \cite{sola2018micro}.
The adjoint matrix of the considered state space $\mathcal{M}$ is given as, \looseness=-1
\begin{gather}
\scalebox{0.8}{
\setcounter{MaxMatrixCols}{20}
$\begin{aligned}
\label{eq:chap:human-motion:ekf-state-adj}
\begin{split}
\gadj{\X_1} =
\begin{bmatrix}
\Rot{}{} & S(\mathbf{p}) \Rot{}{} & \Zero{3}  & \Zero{3} \\
\Zero{3} & \Rot{}{} & \Zero{3}  & \Zero{3} \\
\Zero{3} &  S(\mathbf{v}) \Rot{}{} & \Rot{}{} & \Zero{3} \\
\Zero{3} &  S(\mathbf{d}) \Rot{}{} & \Zero{3} & \Rot{}{}
\end{bmatrix}, \quad \gadj{\X} = \text{blkdiag}(\gadj{\X_1},\ \I{3},\ \Z).
\end{split}
\end{aligned}$
}
\end{gather}
The left Jacobian of the matrix Lie group $\mathcal{M}$ is obtained as a block-diagonal form of the left Jacobians of the constituting matrix Lie groups given as $\gljac{\mathcal{M}}(\err) = \text{blkdiag} \left( \gljac{\SEk{N+2}{3}}\left(\err_1\right), \; \I{3},\ \J\left(\err_{\mathbf{Z}}\right) \right).$
The closed-form expression for the left Jacobian $\gljac{\SEk{N+2}{3}}(.)$ can be found in \cite[Section 2.6]{luo2020geometry}.
The left Jacobian for the translation group of angular velocity is the identity matrix.
\vspace{-0.9em}
\subsection{System Dynamics}
The continuous system dynamics evolving on the group with state $\X \in \mathcal{M}$ is given by the dynamical system, $\frac{d}{dt} \X = f(\X, \mathbf{u}) + \X \ghat{\mathcal{M}}{\textbf{w}}$, where $f$ is the deterministic dynamics, $\mathbf{u} \in \R^m$ is the exogenous control input and  $\textbf{w} \in \mathbb{R}^p$  is a continuous white noise, with covariance $\Q$.
The system dynamics is formulated using rigid body kinematics with constant motion models for the base link along with rigid contact models for the contact points and the foot orientations.
The continuous system dynamics is then given by,
\begin{gather}
\scalebox{0.85}{
\setcounter{MaxMatrixCols}{20}
$\begin{aligned}
	\label{eq:chap:human-motion:ekf-sys-dyn}
	\begin{split}
	\mathbf{\dot{p}} = \mathbf{v}, & \qquad  \mathbf{\dot{R}} = \mathbf{R} \; S(\omega), \\
	\mathbf{\dot{v}} = \mathbf{R} \; (-\noiseLinVel{B}), & \qquad \mathbf{\dot{d}} =  \mathbf{R}\; (-{\Rot{B}{F}}(\encoders)\ \noiseLinVel{F}), \\
	\dot{\omega} =  -\noiseAngVel{B}, & \qquad \mathbf{\dot{Z}} =  \mathbf{Z} \; S(-\noiseAngVel{F}),
	\end{split}
\end{aligned}$
}
\end{gather}
with the left trivialized noise vector defined as,
\begin{gather*}
\scalebox{0.85}{
 $\begin{aligned}
  \label{eq:chap:human-motion:ekf-sys-left-triv-noise}
  & \mathbf{w}  \footnotesize{ \;= \text{vec}\big( \Zeros{3}{1}, \; \Zeros{3}{1}, \; -\noiseLinVel{B}, \;  -{\Rot{B}{F}}(\encoders)\;\noiseLinVel{F}, \; -\noiseAngVel{B},\; -\noiseAngVel{F}\big) },
\end{aligned}$
}
\end{gather*}
and the prediction noise covariance matrix $\mathbf{Q}_c = \expectation{\mathbf{w} \mathbf{w}^T}$. Here, we have used ${\Rot{B}{F}}(\encoders)$ which denotes the rotation of the foot frame with respect to the base link computed through the forward kinematics map using the joint position inputs.
The system dynamics defined in Eq. \eqref{eq:chap:human-motion:ekf-sys-dyn} does not obey the group-affine property \cite[Theorem 1]{barrau2016invariant}. \looseness=-1

\subsubsection{Linearized Error Dynamics}
A right invariant error, $\eta^R = \Xhat \X^{-1}$ is chosen leading to the error dynamics,
\begin{gather}
\scalebox{0.825}{
\setcounter{MaxMatrixCols}{20}
$\begin{aligned}
	\label{eq:chap:human-motion:ekf-sys-errordynamics}
	\begin{split}
	&\frac{d}{dt}(\mathbf{\hat{p}} - \mathbf{\hat{R}}\mathbf{R}^T\;\mathbf{p}) = S(\mathbf{\hat{p}}) \mathbf{\hat{R}}\; \err_\omega + \err_\mathbf{v}, \\
	&\frac{d}{dt}(\mathbf{\hat{R}}\;\mathbf{R}^T) = S(\mathbf{\hat{R}}\; \err_\omega), \\
	&\frac{d}{dt}(\mathbf{\hat{v}} - \mathbf{\hat{R}}\mathbf{R}^T\;\mathbf{v}) = S(\mathbf{\hat{v}}) \mathbf{\hat{R}}\; \err_\omega - \mathbf{\hat{R}}\;\noiseLinVel{B}, \\
	&\frac{d}{dt}(\mathbf{\hat{d}} - \mathbf{\hat{R}}\mathbf{R}^T\;\mathbf{d}) = S(\mathbf{\hat{d}}) \mathbf{\hat{R}}\; \err_\omega - \mathbf{\hat{R}}\;\Rot{B}{F}(\encoders)\;\noiseLinVel{F}, \\
	&\frac{d}{dt}(\hat{\omega} - \omega) = -\noiseAngVel{B},\\
	&\frac{d}{dt}(\mathbf{\hat{Z}}\mathbf{Z}^T) = - S(\mathbf{\hat{Z}}\; \noiseAngVel{F}).
	\end{split}
\end{aligned}$
}
\end{gather}

Using the log-linearity property, the linearized error dynamics and covariance propagation equation then become, \looseness=-1
\begin{gather}
\scalebox{0.85}{
\setcounter{MaxMatrixCols}{20}
$\begin{aligned}
	\label{eq:chap:human-motion:ekf-sys-lin-err-prop}
	\begin{split}
	&\dot{\err} = \mathbf{F}_c\ \err + \gadj{\Xhat}\ \mathbf{w}, \\
	& \dot{\cov}  = \mathbf{F}_c\; \cov + \cov\; \mathbf{F}_c^T + \mathbf{\hat{Q}}_c ,
	\end{split}
\end{aligned}$
}
\end{gather}
where the continuous-time, linearized error propagation matrix $\mathbf{F}_c$ and the prediction noise covariance matrix $\mathbf{\hat{Q}}_c$ are given as, \looseness=-1
\begin{gather}
\scalebox{0.85}{
\setcounter{MaxMatrixCols}{20}
$\begin{aligned}
	\label{eq:chap:human-motion:ekf-err-prop-matrices}
	\begin{split}
	&\mathbf{F}_c\ = \begin{bmatrix}
	\Zero{3} & \Zero{3} & \I{3} & \Zero{3} & S(\mathbf{\hat{p}}) \mathbf{\hat{R}} & \Zero{3} \\
	\Zero{3} & \Zero{3} & \Zero{3} & \Zero{3} & \mathbf{\hat{R}} & \Zero{3} \\
	\Zero{3} & \Zero{3} & \Zero{3} & \Zero{3} & S(\mathbf{\hat{v}}) \mathbf{\hat{R}} & \Zero{3} \\
	\Zero{3} & \Zero{3} & \Zero{3} & \Zero{3} & S(\mathbf{\hat{d}}) \mathbf{\hat{R}} & \Zero{3} \\
	\Zero{3} & \Zero{3} & \Zero{3} & \Zero{3} & \Zero{3} & \Zero{3} \\
	\Zero{3} & \Zero{3} & \Zero{3} & \Zero{3} & \Zero{3} & \Zero{3} \\
	\end{bmatrix}, 
	\mathbf{\hat{Q}}_c  = \gadj{\Xhat}\;\mathbf{Q}_c\;\gadj{\Xhat}^T.
	\end{split}
\end{aligned}$
}
\end{gather}
A discretization of Eq. \eqref{eq:chap:human-motion:ekf-sys-dyn} using a zero-order hold with sampling time $\Delta T$ leads to discrete dynamics, \looseness=-1
\begin{gather}
\scalebox{0.85}{
\setcounter{MaxMatrixCols}{20}
$\begin{aligned}
	\label{eq:chap:human-motion:ekf-disc-dyn}
	\begin{split}
	\mathbf{\hat{p}}_\knext = \; \mathbf{\hat{p}}_\kcurr + \;  \mathbf{\hat{v}}_{\kcurr} \; \Delta T, & \quad  \quad \mathbf{\hat{R}}_{\knext} = \; \mathbf{\hat{R}}_{\kcurr} \; \gexphat{\SO{3}}\big( \hat{\omega}_k \; \Delta T \big), \\
	\mathbf{\hat{v}}_{\knext} = \; \mathbf{\hat{v}}_{\kcurr}, &  \quad \quad \quad \quad \mathbf{\hat{d}}_{\knext} = \; \mathbf{\hat{d}}_{\kcurr},\\
	\hat{\omega}_{\knext} = \; \hat{\omega}_{\kcurr}, & \quad \quad \quad \quad \mathbf{\hat{Z}}_{\knext} = \; \mathbf{\hat{Z}}_{\kcurr}.
	\end{split}
\end{aligned}$
}
\end{gather}
A first-order approximation for the Ricatti equations leads to,
\begin{gather}
\scalebox{0.85}{
\setcounter{MaxMatrixCols}{20}
$\begin{aligned}
	\label{eq:chap:human-motion:ekf-disc-cov-prop}
	\begin{split}
	&\cov_{\knext} = \mathbf{F}_{\kcurr} \; \cov_{\kcurr}\; \mathbf{F}_{\kcurr}^T + \mathbf{Q}_{\kcurr}, \\
	&\mathbf{F}_{\kcurr} = \exp(\mathbf{F}_c \Delta T) \approx  \I{p} + \mathbf{F}_c \Delta T,\\
	&\Q_{\kcurr} = \mathbf{F}_{\kcurr}\; \mathbf{\hat{Q}}_c\; \mathbf{F}_{\kcurr}^T\; \Delta T.
	\end{split}
\end{aligned}$
}
\end{gather}

\subsection{Right Invariant Observations}
We consider measurements having a right-invariant observation structure,  $\mathbf{z}_{\kcurr} = \X_{\kcurr}^{-1}\ \mathbf{b}\ +\ \mathbf{n}_{\kcurr} \in \R^q$ which lead to a time-invariant measurement model Jacobian \cite{barrau2016invariant} and follows the filter update,  \looseness=-1
$\Xhat_{\kcurr}^{+} = \gexphat{\mathcal{M}} \left(\mathbf{K}_\kcurr \left(\Xhat_{\kcurr}\ \mathbf{z}_{\kcurr} - \mathbf{b}\right) \right)\ \Xhat_{\kcurr},$
where, $\mathbf{K}_\kcurr$ is the Kalman gain and $\mathbf{n}_{\kcurr}$ is the noise associated with the observation and $\mathbf{b}$ is a constant vector.
A reduced dimensional gain $\mathbf{K}_\kcurr^r$ can be computed by applying an auxiliary matrix $\Pi$ that selects only the non-zero elements from the full innovation vector in such a way that,
$\mathbf{K}_\kcurr \left(\Xhat_{\kcurr}\ \mathbf{z}_{\kcurr} - \mathbf{b}\right) = \mathbf{K}_\kcurr^r \Pi \Xhat_{\kcurr}\ \mathbf{z}_{\kcurr}.$
The measurement model Jacobian is obtained through a log-linear, first-order approximation of the non-linear error update, \looseness=-1
\begin{gather*}
\scalebox{0.9}{
\setcounter{MaxMatrixCols}{20}
$\begin{aligned}
\errG^{R+}_{\kcurr} = \gexphat{\mathcal{M}} \left(\mathbf{K}_\kcurr \left( \errG^{R}_{\kcurr}\ \mathbf{b}\ -\ \mathbf{b} + \Xhat_{\kcurr}\mathbf{n}_{\kcurr} \ \right) \right) \errG^{R}_{\kcurr}.
\end{aligned}$
}
\end{gather*}

\subsubsection{Relative  candidate  contact point position}
The joint positions $\encoders = \jointPos + \encoderNoise$ obtained from the integration of estimated joint velocities are assumed to be affected by white Gaussian noise $\encoderNoise$ and are used to determine the relative  candidate  contact point positions $h_p(\encoders)$ with respect to the base link using forward kinematics.
The measurement model $h_p(\X)$, right-invariant observation $\mathbf{z}_p$, constant vector $\mathbf{b}_p$, measurement model Jacobian $\mathbf{H}_p$, auxiliary matrix $\Pi_p$ and measurement noise covariance matrix $\mathbf{N}_p$ associated with the relative position measurements are given as, \looseness=-1
\begin{gather}
\scalebox{0.85}{
\setcounter{MaxMatrixCols}{20}
$\begin{aligned}
\label{eq:chap:human-motion:ekf-relposmeas}
	\begin{split}
	&h_p(\X) = \mathbf{\hat{R}}^T(\mathbf{\hat{d}} - \mathbf{\hat{p}}) + \relativeJacobianLeftTrivLinIn{F}{B}(\encoders) \; \encoderNoise \in \mathbb{R}^3, \\
	& \mathbf{z}_p^T = \begin{bmatrix}
	h_p^T(\encoders) & 1 & 0 & -1 & \Zeros{1}{3} & 0 & \Zeros{1}{3}
	\end{bmatrix},\\
	& \mathbf{b}_p^T = \begin{bmatrix}
	\Zeros{1}{3} & 1 & 0 & -1 & \Zeros{1}{3} & 0 & \Zeros{1}{3}
	\end{bmatrix},\\
	& \mathbf{H}_p = \begin{bmatrix}
	-\I{3} & \Zero{3} & \Zero{3} & \I3 & \Zero{3} & \Zero{3}
	\end{bmatrix}, \Pi_p = \begin{bmatrix} \I{3} & \Zeros{3}{10}
\end{bmatrix}\\
	& \mathbf{N}_p = \mathbf{\hat{R}} \; \relativeJacobianLeftTrivLinIn{F}{B}(\encoders) \; \text{Cov}(\encoderNoise) \; \relativeJacobianLeftTrivLinIn{F}{B}^T(\encoders) \; \mathbf{\hat{R}}^T.
	\end{split}
\end{aligned}$
}
\end{gather}

\subsubsection{Zero-velocity Update (ZUPT) aided Left Trivialized Base Velocity}
With a rigid contact assumption, the null velocity of the stance foot can be associated with the base velocity.
The left-trivialized base velocity measurement computed through the joint velocity measurements and the configuration dependent Jacobian of the contact point is used as a right-invariant observation for the filter.
The linear part of the base velocity measurement in the form of $\X_{\kcurr}^{-1}\ \mathbf{b}\ +\ \mathbf{n}_{\kcurr}$, associated with the state variable $\mathbf{v}$ is then described by the quantities, \looseness=-1
\begin{gather}
\scalebox{0.8}{
\setcounter{MaxMatrixCols}{20}
$\begin{aligned}
	\label{eq:chap:human-motion:ekf-ZUPTlin}
	&h_v(\encoders, \encoderSpeeds) = -  \big(\gadj{\Transform{F}{B}}^{-1} \; \relativeJacobianLeftTrivIn{B}{F}(\encoders) \; \encoderSpeeds\big)_{\text{lin}} \in \mathbb{R}^3, \\
	& \mathbf{z}_v^T = \begin{bmatrix}
	h_v(\encoders, \encoderSpeeds) & 0 & -1 & 0 & \Zeros{1}{3} & 0 & \Zeros{1}{3}
	\end{bmatrix},\\
	& \mathbf{b}_v^T = \begin{bmatrix}
	\Zeros{1}{3} & 0 & -1 & 0 & \Zeros{1}{3} & 0 & \Zeros{1}{3}
	\end{bmatrix},\\
	& \mathbf{H}_v = \begin{bmatrix}
	\Zero{3} & \Zero{3} & \I{3} & \Zero{3} & \Zero{3} & \Zero{3} & \Zero{3}
	\end{bmatrix}, \\
	& \Pi_v = \begin{bmatrix} \I{3} & \Zeros{3}{10}
\end{bmatrix}, \mathbf{N}_v = \mathbf{\hat{R}}\text{Cov}(\fkNoiseLinVel{B})\mathbf{\hat{R}}^T,
\end{aligned}$
}
\end{gather}
where, $\gadj{\Transform{F}{B}} \in \R^{6 \times 6}$ is the adjoint transformation transporting the 6D rigid body velocities from the base frame $B$ to the foot frame $F$.
Similarly, the angular part of the left-trivialized base velocity measurement related to the state variable $\omega$ is described by the following quantities for filter update, \looseness=-1
\begin{gather}
\scalebox{0.9}{
\setcounter{MaxMatrixCols}{20}
$\begin{aligned}
\label{eq:chap:human-motion:ekf-ZUPTang}
	\begin{split}
	&h_\omega(\encoders, \encoderSpeeds) = -  \big(\gadj{\Transform{F}{B}}^{-1} \; \relativeJacobianLeftTrivIn{B}{F}(\encoders) \; \encoderSpeeds\big)_{\text{ang}} \in \mathbb{R}^3, \\
	& \mathbf{z}_\omega^T = \begin{bmatrix}
	\Zeros{1}{3} & 0 & 0 & 0 & h_\omega(\encoders, \encoderSpeeds) & -1 & \Zeros{1}{3}
	\end{bmatrix},\\
	& \mathbf{b}_\omega^T = \begin{bmatrix}
	\Zeros{1}{3} & 0 & 0 & 0 & \Zeros{1}{3} & -1 & \Zeros{1}{3}
	\end{bmatrix},\\
	& \mathbf{H}_\omega = \begin{bmatrix}
	\Zero{3} & \Zero{3} & \Zero{3} & \Zero{3} & \I{3} & \Zero{3} & \Zero{3}
	\end{bmatrix}, \\
	& \Pi_\omega = \begin{bmatrix} \Zeros{3}{6} & \I{3} & \Zeros{3}{4}
	\end{bmatrix}, \mathbf{N}_\omega = \text{Cov}(\fkNoiseAngVel{B}).
	\end{split}
\end{aligned}$
}
\end{gather}
We have used $\fkNoiseLinVel{B}$ and $\fkNoiseAngVel{B}$ to denote an additive forward kinematic noise affecting the velocity computations.
\subsection{Left Invariant Observations}
We also consider measurements having a left-invariant observation structure,  $\mathbf{z}_{\kcurr} = \X_{\kcurr}\ \mathbf{b}\ +\ \mathbf{n}_{\kcurr}$ which lead to a time-invariant measurement model Jacobian obtained through a first-order approximation of the error update equation,
\begin{gather*}
\scalebox{0.9}{
\setcounter{MaxMatrixCols}{20}
$\begin{aligned}
\errG^{L+}_{\kcurr} = \errG^{L}_{\kcurr}\ \gexphat{\mathcal{M}} \left(\mathbf{K}_\kcurr \left( \left(\errG^{L}_{\kcurr}\right)^{-1}\mathbf{b}\ -\ \mathbf{b} + \Xhat^{-1}_{\kcurr}\mathbf{n}_{\kcurr} \ \right) \right).
\end{aligned}$
}
\end{gather*}
The filter is then updated by applying the correction through $
\Xhat_{\kcurr}^{+} = \Xhat_{\kcurr}\ \gexphat{\mathcal{M}} \left(\mathbf{K}_\kcurr \left(\Xhat^{-1}_{\kcurr}\ \mathbf{z}_{\kcurr} - \mathbf{b}\right) \right)$.
However, since we have considered right-invariant error $\errG^R$ in our filter design, in order to incorporate the updates from the left-invariant observations, it is necessary to transform the right invariant error to be expressed as the left-invariant error \cite{hartley2020contact} which can be done using the adjoint map as $\err^R =\ \gadj{\Xhat}\ \err^L$ and $\err^L =\ \gadj{{\Xhat^{-1}}}\ \err^R$. 
This implies a switching between the covariance of right- and left-invariant errors as,
\begin{gather}
\scalebox{0.92}{
\setcounter{MaxMatrixCols}{20}
$\begin{aligned}
	\label{eq:chap:human-motion:ekf-cov-switch}
	\begin{split}
	& \cov^L = \gadj{{\Xhat^{-1}}}\ \cov^R\ \gadj{{\Xhat^{-1}}}^T, \\
	& \cov^R = \gadj{\Xhat}\ \cov^L\ \gadj{\Xhat}^T.    
	\end{split}
\end{aligned}$
}
\end{gather}


\subsubsection{Base Collocated Gyroscope}
The gyroscope measurements from the pelvis IMU is used to formulate a left-invariant observation, assuming that they are not affected by any time-varying biases but only by additive white noise $\noiseGyro{B}$.
Considering that the IMU is rigidly attached to the pelvis link and the rotation $\Rot{B}{{B_{\text{IMU}}}}$ between the pelvis link and the pelvis IMU is known, we can compute the left-invariant observations leading to the quantities relevant for filter update as, \looseness=-1
\begin{gather}
\scalebox{0.92}{
\setcounter{MaxMatrixCols}{20}
$\begin{aligned}
\label{eq:chap:human-motion:ekf-basegyro}
	\begin{split}
	&h_g(\tilde{\omega}) = \Rot{B}{{B_{\text{IMU}}}} \yGyro{A}{{B_{\text{IMU}}}} \in \mathbb{R}^3, \\
	& \mathbf{z}_g^T = \begin{bmatrix}
	\Zeros{1}{3} & 0 & 0 & 0 & h_g(\tilde{\omega}) & 1 & \Zeros{1}{3}
	\end{bmatrix},\\
	& \mathbf{b}_g^T = \begin{bmatrix}
	\Zeros{1}{3} & 0 & 0 & 0 & \Zeros{1}{3} & 1 & \Zeros{1}{3}
	\end{bmatrix},\\
	& \mathbf{H}_g = \begin{bmatrix}
	\Zero{3} & \Zero{3} & \Zero{3} & \Zero{3} & \I{3} & \Zero{3} & \Zero{3}
	\end{bmatrix}, \\
	&  \Pi_g = \begin{bmatrix} \Zeros{3}{6} & \I{3} & \Zeros{3}{4}
	\end{bmatrix}, \mathbf{N}_g = \text{Cov}(\noiseGyro{B}) .
	\end{split}
\end{aligned}$
}
\end{gather}
Since, the linearized error update follows standard EKF equations, the gain $\mathbf{K}_\kcurr$ and covariance update can be computed as, \looseness=-1
\begin{gather}
\scalebox{0.85}{
\setcounter{MaxMatrixCols}{20}
$\begin{aligned}
    \begin{split}
        & \mathbf{S}_\kcurr = \mathbf{H}\ \cov_{\kcurr}\ \mathbf{H}^T\ +\ \mathbf{\hat{N}}_\kcurr, \\
        & \mathbf{K}_\kcurr = \cov_{\kcurr}\ \mathbf{H}^T\ \left(\mathbf{S}_\kcurr\right)^{-1}, \\
        & \cov_{\kcurr}^{+} = \left(\I{p} - \mathbf{K}_\kcurr\ \mathbf{H}\right)\ \cov_{\kcurr}.
    \end{split}
\end{aligned}$
}
\end{gather}

\subsection{Non Invariant Observations}
Non-invariant observations considered to be evolving over a distinct matrix Lie group $G^\prime$ are in the form $\Y_\knext = h (\X_\knext) \;\gexphat{G^\prime}(\mathbf{n}_\knext)$ \cite{bourmaud2013discrete}.
The right invariant error leads to the innovation term, \looseness=-1
$
\mathbf{\tilde{z}} = \glogvee{G^\prime}\big(h^{-1}(\Xhat)\; h(\gexp{\mathcal{M}}(-\err) \Xhat)\big),
$ 
and the measurement model Jacobian as, 
$\mathbf{H} = -\frac{\partial}{\partial \err}\; \glogvee{G^\prime}\big(h^{-1}(\Xhat)\; h(\gexp{\mathcal{M}}(-\err)\;\Xhat)\big)\bigg|_{\err = 0}.$

\subsubsection{Relative foot link rotation}
The joint positions $\encoders$ obtained from the integration of estimated joint velocities $\encoderSpeeds$ is passed through the forward kinematics map to determine the relative foot orientations $h_R(\encoders)$. \looseness=-1
\begin{gather}
\scalebox{0.92}{
\setcounter{MaxMatrixCols}{20}
$\begin{aligned}
	\label{eq:eq:chap:human-motion:ekf-relrotmeas}
	\begin{split}
	&h_R(\X) = \mathbf{\hat{R}}^T\;\mathbf{\hat{Z}}\; \gexphat{\SO{3}}\big(\relativeJacobianLeftTrivAngIn{B}{F}(\encoders) \; \encoderNoise\big) \in \SO{3}, \\
	& \mathbf{H}_R = \begin{bmatrix}
	\Zero{3} & -\mathbf{\hat{Z}}^T & \Zero{3} & \Zero{3} & \Zero{3} & \mathbf{\hat{Z}}^T
	\end{bmatrix}, \\
	& \mathbf{N}_R = \; \relativeJacobianLeftTrivAng{B}{F}(\encoders) \; \text{Cov}(\encoderNoise) \; {\relativeJacobianLeftTrivAng{B}{F}}^T(\encoders).
	\end{split}
\end{aligned}$
}
\end{gather}

\subsubsection{Terrain Height Update}
For candidate points that are actively in contact with the environment, the height measurement from a known map affected by noise $\mathbf{w}_\mathbf{d}$ is used to update the filter states.
High covariance values are associated with $(d_x, d_y)$ coordinates while map covariance values is set for the height $d_z$.
The measurement update equations for non-invariant observation can be written as,
\begin{gather}
\scalebox{0.92}{
\setcounter{MaxMatrixCols}{20}
$\begin{aligned}
	\label{eq:eq:chap:human-motion:ekf-relrotmeas}
	\begin{split}
	&h_d(\X) =\mathbf{d} + \mathbf{w}_\mathbf{d} \in \T{3}, \\
	& \mathbf{H}_d = \begin{bmatrix}
	\Zero{3} & S(\mathbf{\hat{d}}) & \Zero{3} & -\I{3} & \Zero{3} & \Zero{3}
	\end{bmatrix}, \\
	& \mathbf{N}_d = \; \text{Cov}(\mathbf{w}_d).
	\end{split}
\end{aligned}$
}
\end{gather}

\subsubsection{Contact Plane Orientation Update}
In cases where, the foot is in planar rigid contact with the environment, we enable a contact plane orientation update.
We relate the contact plane orientation  measurement affected by noise $\mathbf{w}_c$ directly to the foot orientation $\Z$. The filter update quantities associated with this measurement can be summarized as, \looseness=-1
\begin{gather}
\scalebox{0.92}{
\setcounter{MaxMatrixCols}{20}
$\begin{aligned}
	\label{eq:eq:chap:human-motion:ekf-relrotmeas}
	\begin{split}
	&h_c(\X) = \mathbf{\hat{Z}}\; \gexphat{\SO{3}}(\mathbf{w}_c) \in \SO{3}, \\
	& \mathbf{H}_c = \begin{bmatrix}
	\Zero{3} & \Zero{3} & \Zero{3} & \Zero{3} & \Zero{3} & \mathbf{\hat{Z}}^T
	\end{bmatrix}, \\
	& \mathbf{N}_c = \; \text{Cov}(\mathbf{w}_c).
	\end{split}
\end{aligned}$
}
\end{gather}

The state update in the tangent space becomes $\mathbf{m}^{-}_\knext = \mathbf{K}_\knext\; \mathbf{\tilde{z}}_\knext $, which is used for the state reparametrization $\Xhat_\knext = \gexphat{\mathcal{M}}\big(\mathbf{m}^{-}_\knext\big) \Xhat_\kpred$. 
The state covariance is updated as $\cov_\knext = \J^l_\mathcal{M}(\mathbf{m}^{-}_\knext)\;(\I{p} - \mathbf{K}_\knext\;\mathbf{H}_\knext)\cov_\kpred\;\J^l_\mathcal{M}(\mathbf{m}^{-}_\knext)^T$.


\section{Experiments and Results}
\label{sec:RESULTS}
In this section, we present the results from the experimental evaluation of the proposed estimation algorithm for reconstructing the whole body kinematic motion of the human performing walking-in-place and free walking motions.

\subsection{Experimental Setup}
The experiments were conducted with a human subject wearing sensorized suit and shoes developed by \emph{iFeel Tech}\footnote{\url{https://ifeeltech.eu/}} which are part of a whole-body wearable technology aimed at real-time human motion tracking and articular stress analysis.
Each shoe worn by the human is equipped with 6-axis force-torque sensors at its front and rear part of the sole and streams a ground reaction wrench measurement at $60 \si{Hz}$.
The sensorized suit consists of 10 \emph{iFeel nodes} distributed as follows: 2 nodes on each arm (upper arm, fore arm), 2 nodes on each leg (upper leg, lower leg), one each on the Pelvis and stern and 2 nodes on the foot as part of the shoes. 
Each iFeel node has a Bosch BNO055 IMU that  streams measurements at $60 \si{Hz}$. \looseness=-1 

For the articulated rigid-body human model, we use the URDF model relevant to the subject available from the \emph{human-gazebo}\footnote{\url{https://github.com/robotology/human-gazebo}} project. 
A subset of 29 Degrees of Freedom (DoFs), from a total of 48 DoFs, related to the limb and torso kinematics is used to reconstruct a coarse full-body motion. \looseness=-1

For a validation of the estimated floating base pose, we perform a comparison with a base trajectory obtained from two different methods. 
The first method simply uses a \emph{Vive}\footnote{\url{https://www.vive.com/us/accessory/tracker3/}} motion tracker that is mounted on the base link of the human subject (\emph{Pelvis}) in order to track the base pose trajectory.
Another base trajectory is obtained from a base line algorithm, referred to as HSP, which relies solely on the dynamical IK, circumventing the connection with the proposed EKF.
This approach uses the known floor position as a pseudo-target measurement within the IK whenever a foot contact is detected, thereby constraining the whole body motion to achieve legged odometry. \looseness=-1

The data from the \emph{iFeel} suit and shoes along with the tracker pose are logged at a frequency of $50 \si{Hz}$ during the experiments and is played back in real-time to run the Whole Body Kinematics (WBK) estimation and the baseline algorithm (HSP) at $50 \si{Hz}$.
The output trajectories are then saved in Matlab data format and aligned to the same reference frame for plotting the results.
The trajectories are aligned using $\A \X = \Y \B$ calibration method to find an unknown transformation between the tracker and Pelvis and another between the world frames. \looseness=-1

\subsection{Experiments}
For the experiments, the subject initially stays in an extended arm configuration, usually referred to as \emph{TPose} configuration before performing the actual motion. A calibration procedure is performed during \emph{TPose} to align the measurements from different IMUs to a common reference frame for the estimation.
We make use of only absolute orientation measurements from the IMUs as inputs to the dynamical IK block.
The differential inverse kinematics is solved using a QP solver with the consideration of joint limits for the legs tuned to allow for physically feasible motion.
The rectangular foot geometry is set using the dimensions of the sensorized shoes.
Known constant floor height is passed as a measurement to the EKF every time a candidate point on the foot is detected to be in contact.
When all the candidate points on the foot are in contact, a contact orientation reflecting the flat planar contact surface is passed as measurement by setting the tilt angles to zero.
The estimated base pose is not fed back to the Dynamical IK block. 
This allows the joint state estimation to be invariant of the floating base pose and depends only on the orientation measurements and relative kinematics.
The accuracy in aligning the estimated trajectories with the Vive tracker trajectory is low due to missing data in the latter, and for this reason we pertain to a qualitative comparison for discussion. \looseness=-1

\subsubsection{In-Place Walking Motion}
\begin{figure}[!t]
\centering
\includegraphics[width=0.5\textwidth]{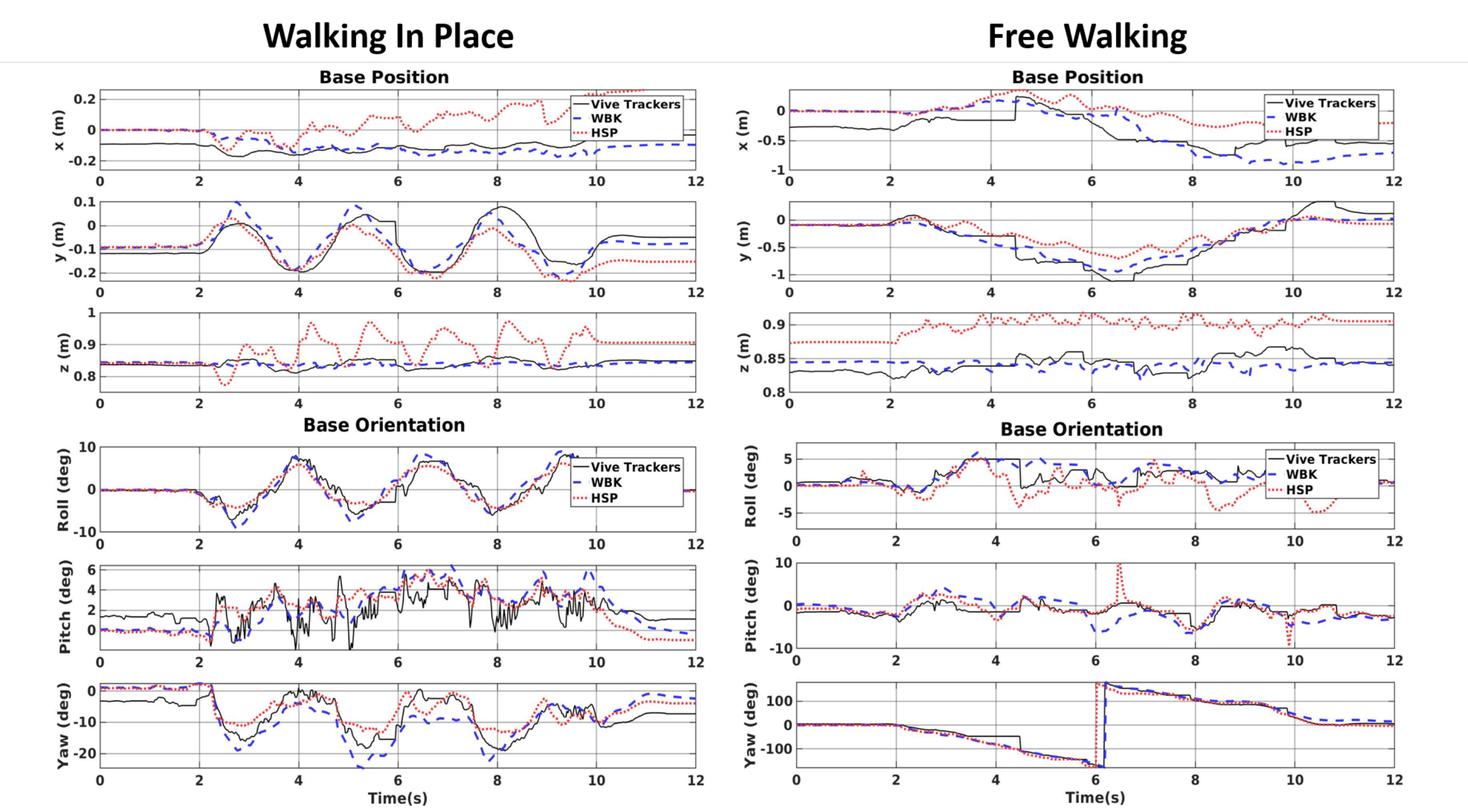}
\caption{Evolution of base pose trajectories estimated by proposed WBK (blue dashed), base line HSP (red dotted), and that  measured from Vive trackers (black solid) for a walking-in-place (left) and free walking motions (right). \looseness=-1}
\vspace{-1.25em}
\label{fig:plots}
\end{figure}

Figure \ref{fig:walk-in-place} shows one snapshot from the walking-in-place trajectory performed by the human and the estimated trajectory at that time instant.
The overall base motion is fairly reconstructed in comparison with the Vive tracker trajectory as seen in Figure \ref{fig:plots}. 
The baseline algorithm HSP is seen to suffer considerably more from drifts in the forward direction $\mathbf{x}$ and along the base height $\mathbf{z}$ leading to resemble a stair-case walking motion. 
WBK suffers from low position drifts owing to the contact-aided measurements such as terrain height and contact plane updates. 
WBK is seen to be sensitive to the covariance of the base gyroscope measurements leading to varying degree of errors in the roll and heading angles of the base orientation. 
A considerable tuning effort is required to obtain reliable pose estimates from the EKF.
\looseness=-1
\begin{figure*}[t!]
    \centering
    \includegraphics[height=72mm, width=0.9\textwidth]{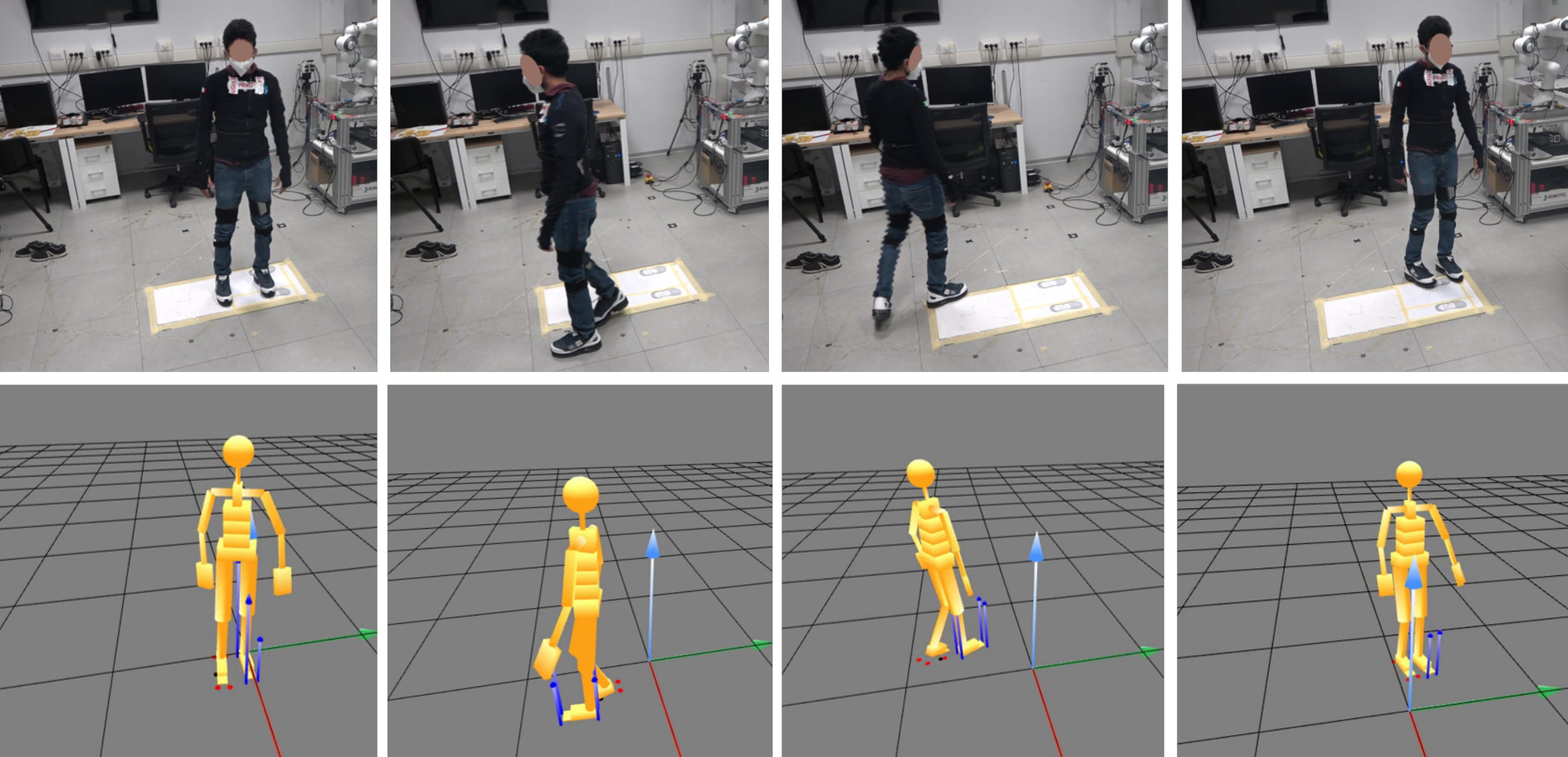}
    \caption{Snapshots from a free-walking experiment performed by the subject (top) and corresponding motion reconstruction from the proposed WBK estimation (bottom). Blue arrows represent contact normal forces for points in contact with the environment, while red points indicate loss of contact, and the black dot represents the local center of pressure which retains the last seen position when exiting the support polygon.}
    \vspace{-1.25em}
    \label{fig:walk}
\end{figure*}

\begin{figure}[!t]
\centering
\includegraphics[scale=0.15, width=0.3\textwidth]{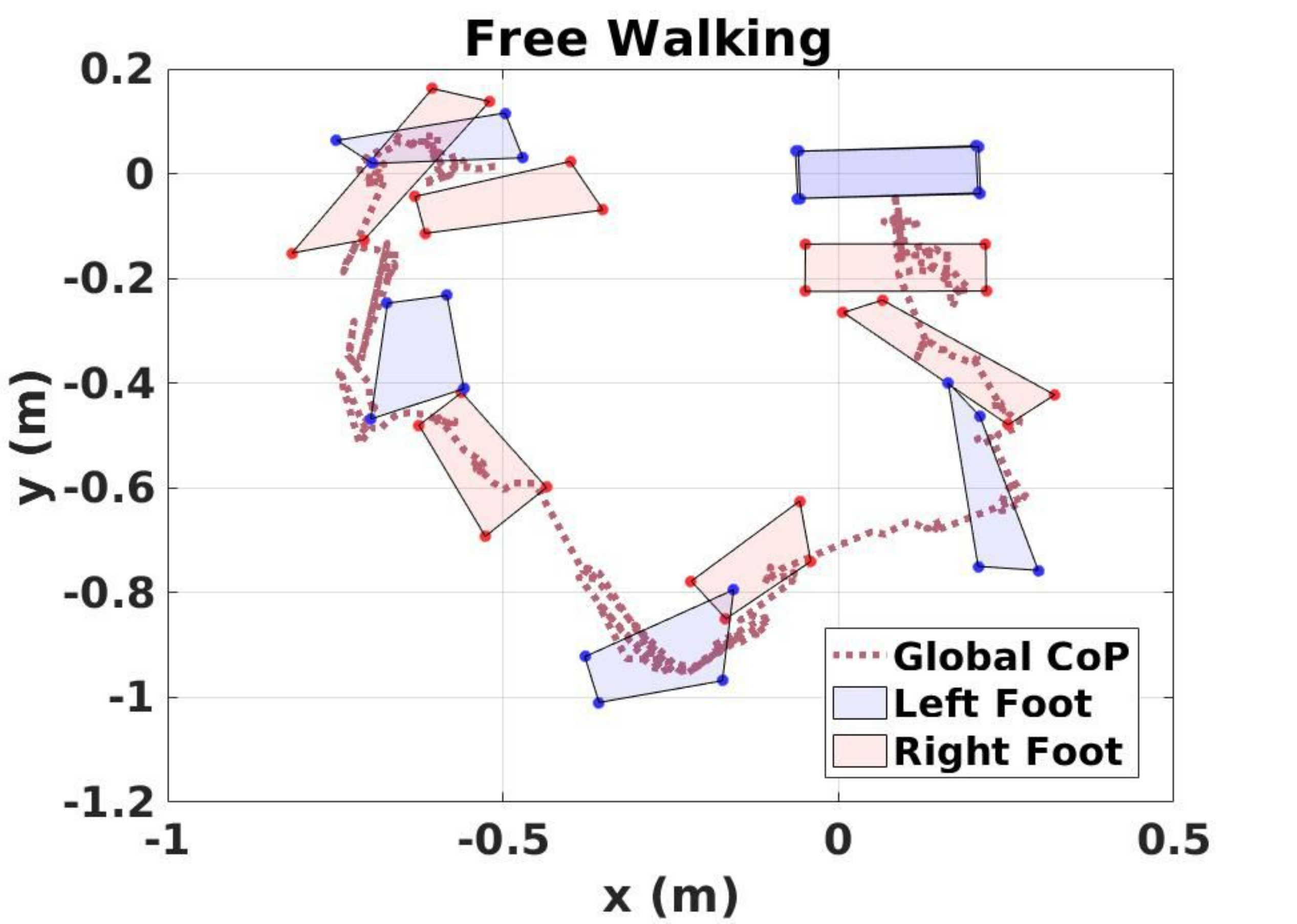}
\caption{XY-position of the global CoP trajectory computed using WBK  estimated base pose along with estimated position of contact points on the feet at first instants of full planar foot with the ground.}
\vspace{-2em}
\label{fig:cop}
\end{figure}

\subsubsection{Free Walking Motion}

Figure \ref{fig:walk} shows snapshots from the walking trajectory performed by the human and the estimated trajectory at those time instants.
The overall walking motion seems to reasonably reconstructed and from Figure \ref{fig:plots}, it can be seen that WBK performs comparably with the HSP algorithm for a walking motion.
HSP gains $5 \si{\centi\meter}$ on the base height for a $12 \si{\second}$ walking over a flat surface, while WBK maintains the nominal base height during walking with deviations less than $1 \si{\centi\meter}.$ 
WBK is seen to suffer more in the forward walking and heading directions, eventhough it can be seen from Figure \ref{fig:cop} that the evolution of the global CoP computed from the estimated base pose is consistent with the walking motion.
The CoP moves from the heel to toe for a forward walking motion and changes sharply during turns while always remaining bounded between the feet polygons obtained from the estimated positions of candidate contact points on the feet.
The rectangular shape of the feet is not preserved by the estimated positions on the XY-plane eventhough the contact-aided measurement updates are enforced.
This is because the knowledge of the rectangular feet is not exploited in the EKF and the contact position states are completely decoupled from the feet orientation states and are related only to the base orientation.
The filter can be constrained to obtain better estimates by introducing this information. \looseness=-1
\section{Conclusion}
\label{sec:CONCLUSION}
In this paper, we presented an approach for whole body kinematics estimation of a human using distributed inertial and force-torque sensing, in the absence of position sensors.
This is done by extending a Dynamical IK approach for joint state estimation with floating base estimation using contact-aided filtering on Lie groups. 
The proposed method was validated qualitatively for flat surface walking and in-place walking motions in comparison with trajectories obtained from Vive motion tracking system. \looseness=-1

There is definitely room for many improvements to the proposed algorithm especially concerning the prediction model and thresholding based contact inference.
Nevertheless, the methodology allows for a modular and effective approach for full body estimation by exploiting filtering on Lie groups in cascade with inverse kinematics.
Further, the method is directly applicable also to robots with similar sensing capabilities thereby allowing for a unified kinematic estimation approach for both humans and robots.

Future work aims to extend the EKF with a more reliable prediction model that is learned from data along with relevant learned measurement models that will allow removing the dependence on using absolute orientation measurements which are usually affected by magnetometer drifts.
\vspace{-0.5em}
\addtolength{\textheight}{-0.18cm}   
\bibliography{bibliography}
\bibliographystyle{IEEEtran}


\end{document}